\documentclass[12pt]{article}

\usepackage[utf8]{inputenc}
\usepackage[T1]{fontenc}
\usepackage{hyperref}
\usepackage{url}
\usepackage{booktabs}
\usepackage{amsfonts}
\usepackage{microtype}
\usepackage{graphicx}
\usepackage{float}  % For better control of float positioning
\usepackage{etoolbox}

\AfterEndEnvironment{table}{\vspace{1em}}
\graphicspath{{./figures/}}

\usepackage[margin=1in]{geometry}

\usepackage{titlesec}
\titleformat{\section}{\normalfont\Large\bfseries}{\thesection}{1em}{}
\titleformat{\subsection}{\normalfont\large\bfseries}{\thesubsection}{1em}{}
\titlespacing*{\section}{0pt}{2em}{1em}
\titlespacing*{\subsection}{0pt}{1.5em}{0.5em}

\usepackage{setspace}
\title{\Large QuaLLM-Health: An Adaptation of an LLM-Based Framework for Quantitative Data Extraction from Online Health Discussions}

\author{
\normalsize
Ramez Kouzy$^{1,}$\footnote{\href{mailto:rkouzy@mdanderson.org}{rkouzy@mdanderson.org}} , Roxanna Attar-Olyaee$^{2,*}$, Michael K. Rooney$^1$, Comron J. Hassanzadeh$^1$,\\ \normalsize Junyi Jessy Li$^3$, Osama Mohamad$^1$\\
\\
\small$^1$The University of Texas MD Anderson Cancer Center, Houston, Texas, USA\\
\small$^2$Texas Tech Health Science Center El Paso, El Paso, Texas, USA\\
\small$^3$University of Texas at Austin, Austin, Texas, USA\\[0.5em]
{\small *co-first authors}
}
\begin{document}

\maketitle
\begin{abstract}
\textbf{Background:} Health-related discussions on social media like Reddit offer valuable insights, but extracting quantitative data from unstructured text is challenging. In this work, we present an adapted framework from QuaLLM into \textbf{QuaLLM-Health} for extracting clinically relevant quantitative data from Reddit discussions about glucagon-like peptide-1 (GLP-1) receptor agonists using large language models (LLMs).

\textbf{Methods:} We collected 410,710 posts and comments from five GLP-1–related Reddit communities using the Reddit API in July 2024. After filtering for cancer-related discussions with predefined keywords, 2,059 unique entries remained. We developed annotation guidelines to manually extract variables such as cancer survivorship, family cancer history, cancer types mentioned, risk perceptions, and discussions with physicians. Two domain-expert annotators independently annotated a random sample of 100 entries to create a gold-standard dataset, achieving high inter-annotator agreement (Fleiss' kappa $\kappa \geq 0.8$) for key variables. We then employed iterative prompt engineering with OpenAI's GPT-4o-mini on the gold-standard dataset to build an optimized pipeline that allowed us to extract variables from dataset in question.

\textbf{Results:} The optimized LLM achieved accuracies above 0.85 for all variables, with precision, recall and F1 score macro averaged > 0.90, indicating balanced performance. Stability testing showed a 95\% match rate across runs, confirming consistency. Applying the framework to the full dataset enabled efficient extraction of variables necessary for downstream analysis, costing under \$3 and completing in approximately one hour.

\textbf{Conclusion:} QuaLLM-Health demonstrates that LLMs can effectively and efficiently extract clinically relevant quantitative data from unstructured social media content. Incorporating human expertise and iterative prompt refinement ensures accuracy and reliability. This methodology can be adapted for large-scale analysis of patient-generated data across various health domains, facilitating valuable insights for healthcare research.
\end{abstract}

\section{Introduction}
The increasing volume of health-related discussions on social media platforms, such as Reddit, presents a unique opportunity to derive meaningful insights about patient experiences, medication effects, and public health concerns.\href{https://paperpile.com/c/crHGMz/9WH1}{\textsuperscript{1}} However, transforming these unstructured discussions into actionable data poses significant challenges. In this study, we present QuaLLM-Health, an adaptation of the QuaLLM framework specifically designed to extract clinically relevant quantitative data from Reddit conversations about glucagon-like peptide-1 (GLP-1) receptor agonists.\href{https://paperpile.com/c/crHGMz/tG8Y}{\textsuperscript{2}} This approach integrates structured prompt engineering, human-in-the-loop validation, and a comprehensive evaluation methodology to effectively translate social media content into a rich source of information for healthcare research.

Our goal with QuaLLM-Health is to leverage the strengths of large language models (LLMs) for the analysis of unstructured online forum data, enabling the extraction of actionable quantitative insights in an efficient and cost-effective manner. By integrating expert knowledge throughout the process, this framework ensures accuracy, consistency, and applicability to the healthcare domain, offering a blueprint for future research in health informatics and patient-centered studies.

\subsection{Overview of the Framework}
Our adapted framework encompasses the following key components (Figure 1):

\begin{enumerate}
    \item \textbf{Data Collection}: Systematic gathering of relevant Reddit posts and comments.

    \item \textbf{Data Preprocessing}: Rigorous filtering and cleaning of the collected data based on established criteria.

    \item \textbf{Annotation Guideline Development}: Creation of a detailed guideline for consistent variable extraction.

    \item \textbf{Human Annotation}: Generation of a high-quality, consensus-based gold standard dataset.

    \item \textbf{LLM Prompt Engineering}: Implementation of iterative prompt engineering techniques with a LLM for automated extraction.

    \item \textbf{Evaluation}: Thorough assessment of the LLM’s performance using the gold standard dataset.
    
    \item \textbf{Execution}: Deployment of a fine-tuned pipeline on a large dataset for quantitative data extraction for downstream analysis.
    \end{enumerate}

\begin{figure}[ht]
    \centering
    \includegraphics[width=\textwidth]{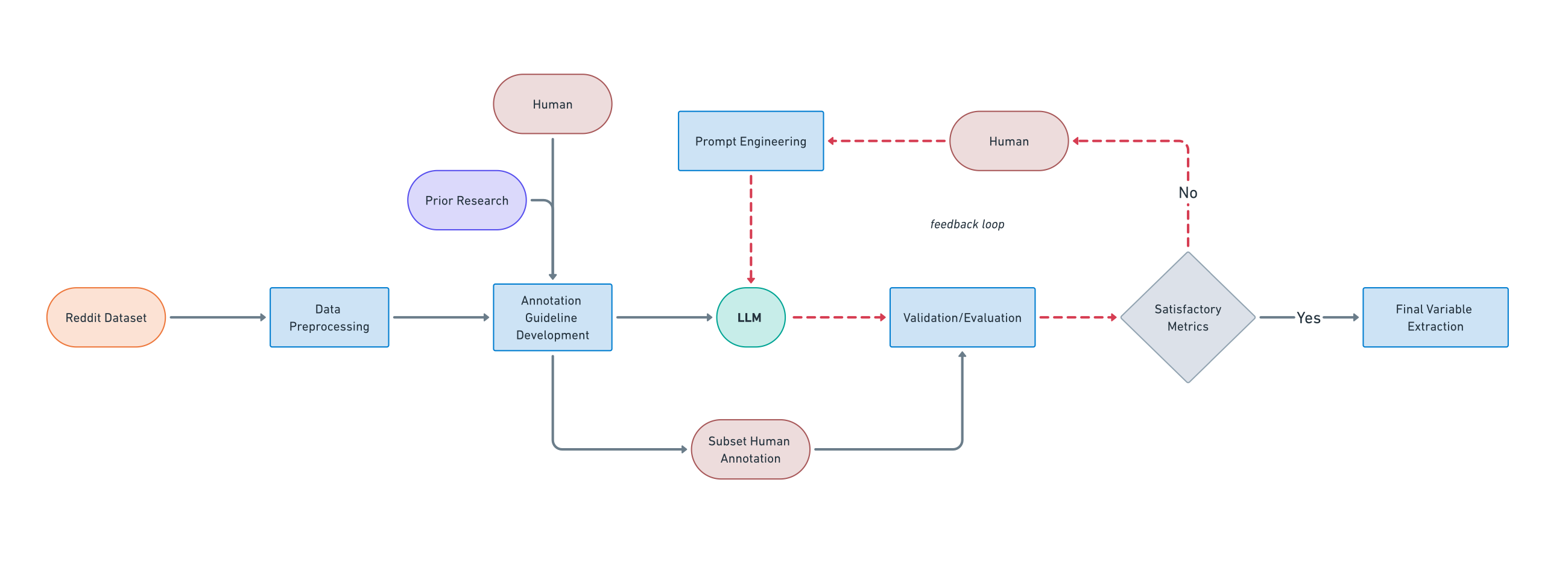}
    \caption{Overview of the QuaLLM-Health framework showing the pipeline from data collection through execution.}
    \label{fig:framework}
\end{figure}

\section{Data Collection and Preprocessing}

\subsection{Data Collection}
We initiated the data collection process by identifying five Reddit communities (subreddits) with the highest user engagement related to GLP-1 receptor agonists. These included subreddits focused on the medication names (r/Semaglutide, r/WegovyWeightLoss, r/Zepbound, r/Ozempic, and r/Mounjaro). Data were collected using the Reddit API in July 2024.\href{https://paperpile.com/c/crHGMz/59xf}{\textsuperscript{3}} This initial data collection resulted in approximately 410,710 entries, comprising both posts and comments.

\subsection{Data Filtering}
We developed a regex pipeline to automate the filtering process, using keywords and phrases specific to GLP-1 receptor agonists as well as cancer-related terms to identify relevant entries. The cancer-related terms were identified using a predefined list including 'cancer', 'malignancy', 'tumor', 'neoplasm', 'carcinoma', 'sarcoma', 'leukemia', 'lymphoma', 'metastasis', 'oncology', 'biopsy', 'chemotherapy', 'radiation therapy', 'malignant', and 'benign tumor'. This filtering step reduced the dataset to approximately 2,390 entries.

\subsection{Data Cleaning and Preparation}
We removed duplicates and non-English posts, which left us with a dataset comprised approximately 2,059 unique entries suitable for analysis. We then selected a random sample of 100 entries from the filtered dataset, which were given to human annotators for validation and to assist in prompt engineering.

\section{Annotation Guideline Development}
We developed an annotation guideline describing variables that could be extracted as binary or categorical variables. This was used to label the textual entries (posts or comments) in a more consistent and clear way. This approach aimed to maintain consistency and clarity in identifying clinically relevant information. The annotation guideline, data used, and code are available at \href{https://github.com/ramezkouzy/GLP1-LLM}{https://github.com/ramezkouzy/GLP1-LLM}.  The guideline included the following key areas for annotation:

\begin{itemize}
    \item \textbf{Inclusion and Exclusion Criteria}: Posts were categorized based on whether they discussed GLP-1 receptor agonists in the context of cancer. Instances that did not meet the inclusion criteria were labeled with specific exclusion reasons to maintain dataset relevance and clarity.
    \item \textbf{Cancer Survivorship and Related Details}: We identified mentions of cancer survivors, including whether they were currently taking GLP-1 medications and if they are taking it to lose weight.
    \item \textbf{Family Cancer History and Cancer Type}: Posts mentioning a family history of cancer or specifying types of cancer were annotated to facilitate deeper analysis of user backgrounds and cancer types involved. Unusual or unlisted types of cancer were captured with a free-text entry.
    \item \textbf{Cancer Risk and Discussions with Physicians}: Mentions related to cancer risk, including concerns, discussions of increased or decreased risk, and whether these concerns were addressed with healthcare professionals.
    \item \textbf{Timing of Cancer Diagnosis}: Posts mentioning a cancer diagnosis that occurred after starting GLP-1 medication were flagged to help identify any temporal relationships discussed by users.
    \item \textbf{Information Seeking}: Instances where users explicitly sought information regarding cancer risks associated with GLP-1 medications were annotated to understand user awareness and areas where more education might be needed.
\end{itemize}

This systematic and structured annotation allowed us to create a high-quality, gold-standard dataset that could reliably be used to evaluate LLM performance and guide prompt engineering. The finalized annotation guideline is attached in the published github library.\footnote{\url{https://github.com/ramezkouzy/GLP1-LLM}}.

\section{Human Annotation Process}
Two annotators with domain expertise were tasked with independently annotating the same random sample of 100 entries from the cleaned dataset. Annotations were recorded in a standardized format, capturing all variables as per the guideline. Post-annotation, discrepancies were adjudicated by discussion. The variables and the frequencies are presented in Table 1. Inter-annotator agreement was quantified using Fleiss' kappa, with results for each variable as follows:

\begin{itemize}
    \item \textbf{Variables with High Agreement ($\kappa \geq 0.8$)}: Discussions about GLP-1 receptor agonists in the context of cancer, mentions of being a cancer survivor, types of cancer mentioned, cases where a cancer survivor is also taking GLP-1 medication, mentions of family cancer history, discussions of cancer risk with a physician, mentions of other cancer types not previously listed, and discussions about GLP-1 potentially decreasing cancer risk. These variables showed strong consensus among annotators, indicating reliable extraction for these key areas.
    \item \textbf{Variables with Moderate Agreement ($0.6 \leq \kappa < 0.8$)}: Mentions of weight loss among cancer survivors, cancer diagnosis after starting GLP-1 medication, seeking information about cancer risks, mentions of increased cancer risk, and expressions of concern regarding cancer risk. These variables had moderate agreement, suggesting a reasonable level of consistency but room for refinement in the annotation guidelines or further training.
    \item \textbf{Variables with Low Agreement ($\kappa < 0.6$)}: Ability to assess misinformation, overall sentiment of the post, tone of the discussion, general context, and references to misinformation. These variables showed significant variability in interpretation between annotators and were excluded from LLM performance evaluations, though they may provide useful insights in qualitative analyses.
\end{itemize}

\begin{table} [h!]
  \caption{Cancer-related variables in the random double-annotated sample (N=100).}
  \centering
  \begin{tabular}{lc}
    \toprule
    Variable & N \\
    \midrule
    Cancer Type & \\
    \quad Thyroid Cancer & 44 \\
    \quad Breast Cancer & 8 \\
    \quad Gyn Cancer & 2 \\
    \quad Pancreatic Cancer & 2 \\
    \quad Other & 8 \\
    \quad No Type Mentioned & 32 \\
    \midrule
    Cancer History and Experience & \\
    \quad Is Cancer Survivor & 33 \\
    \quad Survivor Taking GLP-1 & 26 \\
    \quad Cancer Diagnosis After Medication & 13 \\
    \midrule
    Risk Communication and Perception & \\
    \quad Mentions Cancer Risk & 57 \\
    \quad Concerned About Cancer Risk & 29 \\
    \quad Seeking Cancer Risk Data & 9 \\
    \quad Discussed Risk with Physician & 17 \\
    \quad GLP-1 Decreasing Cancer Risk & 13 \\
    \bottomrule
  \end{tabular}
  \label{tab:cancer_frequencies}
\end{table}

\section{LLM Prompt Engineering and Evaluation}

\subsection{Initial Prompting Strategy}
We utilized Open AI's gpt-4o-mini-2024-07-18 via API calls, providing it with a carefully structured prompt. This prompt was based on the annotation guidelines that human annotators followed, ensuring that the language model received explicit instructions for each variable. Alongside this, we also provided a JSON schema that defined the expected output format, including specific fields and data types. Our goal with this zero-shot prompting strategy was to evaluate the model's baseline ability to understand and extract variables without relying on any prior examples. The initial evaluation of the LLM's output against the gold standard annotations revealed mixed results, which are presented in Table 2. The model accurately identified basic variables but struggled with more nuanced aspects, such as detecting cancer type and risk discussion detection.

\begin{table} [h!]
  \caption{Performance metrics at baseline.}
  \centering
  \begin{tabular}{lcccc}
    \toprule
    Category & Precision & Recall & F1 \\
    \midrule
    inclusion & 0.943 & 0.510 & 0.640 \\
    exclusion\_reason  & 1.000 & 0.500 & 0.667 \\
    is\_survivor & 0.939 & 0.938 & 0.936 \\
    is\_survivor\_and\_taking\_med & 0.886 & 0.865 & 0.848 \\
    cancer\_type & 0.603 & 0.510 & 0.539 \\
    other\_cancer\_type & 0.936 & 0.823 & 0.875 \\
    is\_survivor\_weight\_loss & 0.844 & 0.833 & 0.806 \\
    cancer\_diagnosis\_after\_medication & 0.910 & 0.917 & 0.910 \\
    mentions\_cancer\_risk & 0.771 & 0.740 & 0.741 \\
    concerned\_about\_cancer\_risk & 0.812 & 0.812 & 0.812 \\
    seeking\_cancer\_risk\_data & 0.903 & 0.917 & 0.906 \\
    discussed\_risk\_with\_physician & 0.921 & 0.906 & 0.911 \\
    discussion\_GLP1\_decreasing\_cancer\_risk & 0.889 & 0.885 & 0.851 \\
    \midrule
    Macro-average & 0.874 & 0.781 & 0.803 \\ 
    \bottomrule
  \end{tabular}
  \label{tab:performance_metrics}
\end{table}

\subsection{Prompt Engineering}
To address these limitations, we employed iterative improvements, including chain-of-thought reasoning, few-shot prompting, and edge case inclusion.\href{https://paperpile.com/c/crHGMz/DIhW}{\textsuperscript{4}} The prompts used are detailed in the github library.\footnote{\url{https://github.com/ramezkouzy/GLP1-LLM}} Importantly, we did not use examples directly from the evaluation dataset. We employed a human-in-the-loop process, as depicted in Figure 1, to fine-tune the prompts and create new examples that specifically highlighted edge cases where the model struggled. We also set the temperature at 0.0 to decrease the stochasticity of the responses, maximizing the consistency of the classification. Additionally, we utilized the model's reasoning to identify discrepancies and refine the prompts further, ultimately enhancing alignment with the gold standard dataset and reducing inconsistencies.

Each iteration of prompt refinement was followed by a re-evaluation of the LLM’s performance metrics, allowing us to progressively enhance the model's extraction capabilities and ensure alignment with our gold standard dataset. The final LLM configuration achieved accuracy of at least 0.85 across all included variables, demonstrating reliable extraction performance. Precision, recall, and F1 macro average were higher than 0.90 indicating a well-balanced performance across all variables. Detailed performance metrics for each variable are presented in Table 3.

\begin{table} [h!]
  \caption{Performance metrics after prompt engineering.}
  \centering
  \begin{tabular}{lccc}
    \toprule
    Category & Precision & Recall & F1 \\
    \midrule
    inclusion & 0.944 & 0.950 & 0.947 \\
    exclusion\_reason & 1.000 & 0.979 & 0.989 \\
    is\_survivor & 0.921 & 0.917 & 0.914 \\
    is\_survivor\_and\_taking\_med & 0.886 & 0.865 & 0.848 \\
    cancer\_type & 0.920 & 0.906 & 0.906 \\
    other\_cancer\_type & 0.929 & 0.927 & 0.925 \\
    is\_survivor\_weight\_loss & 0.883 & 0.885 & 0.881 \\
    cancer\_diagnosis\_after\_medication & 0.924 & 0.927 & 0.919 \\
    mentions\_cancer\_risk & 0.822 & 0.823 & 0.822 \\
    concerned\_about\_cancer\_risk & 0.851 & 0.854 & 0.851 \\
    seeking\_cancer\_risk\_data & 0.971 & 0.969 & 0.969 \\
    discussed\_risk\_with\_physician & 0.916 & 0.896 & 0.902 \\
    discussion\_GLP1\_decreasing\_cancer\_risk & 0.878 & 0.896 & 0.881 \\
    \midrule
    Macro Average & 0.911 & 0.909 & 0.904 \\
    \bottomrule
  \end{tabular}
  \label{tab:performance_metrics_with_macro}
\end{table}

\subsection{Stability Testing}
To evaluate the consistency of the LLM, we ran the final prompt configuration five times on the gold standard dataset. The results showed an average pairwise match rate of 95\% across runs, indicating strong agreement. Variables with clear textual cues demonstrated high consistency, while minor variations were observed in more complex variables that required inference. Overall, these findings confirm the LLM's stability in extracting variables under the optimized prompt settings.

\section{Application of the Framework}
We applied the optimized prompt to the full dataset of 2,059 unique entries. The LLM successfully extracted variables required for downstream analysis in an efficient manner, with the entire process—including prompt engineering—costing under \$3 and taking around one hour to complete. The results will be presented in future analysis.

\section{Discussion}
In this report, we present an adaptation of the QuaLLM framework that leverages the power of LLMs to efficiently analyze large volumes of unstructured text data, with a particular focus on extracting quantitative data. This adapted framework demonstrates value for similar and broader use cases where both quantitative and structured data are required for research purposes.

In this adaptation, we build on previously described concepts, enhancing the utility of LLMs for use cases such as healthcare-related research. We employed iterative prompting, which has proven to be highly effective, as well as few-shot prompting, both of which improved model performance.\href{https://paperpile.com/c/crHGMz/SpaU+7zDG}{\textsuperscript{5,6}} We also emphasize the importance of involving humans at every stage of the workflow, especially in downstream deployment. Insights from domain experts play a crucial role in maximizing performance by addressing domain-specific nuances and challenges.

In addition to enhancing LLM utility, our framework exemplifies a cost-effective and time-efficient method for large-scale text analysis. The structured approach adopted in this study underscores the potential of LLMs in producing reliable and reproducible results even in complex domains such as healthcare. The combination of human expertise and iterative model tuning facilitated a nuanced understanding of social media data, which is often noisy and context-specific. This methodology not only highlights the adaptability of LLMs but also serves as a guide for future research that aims to leverage large language models for extracting actionable insights from unstructured datasets. Moreover, this approach saves significant resources and personnel time, reducing the need for manual data extraction and analysis. Future iterations of this framework could allow even more researchers to find and adapt the methodology to address their specific use cases, further expanding the potential applications across various domains.

\subsection{Limitations}
We acknowledge that the generalizability of our human evaluation is limited by the relatively small sample size and its focus on specific pipeline stages. More sophisticated fine-tuned or larger models might perform better, but our approach prioritizes efficiency and cost-effectiveness. Setting up the pipeline took considerable time, but we aimed to balance efficiency with accuracy and cost. Even with high stability, the model may still exhibit slight variations between runs, which could make it unsuitable for high-stakes applications like clinical deployment. However, it is also important to remember that human evaluators are not fault-proof, and the comparator is often a human with their own limitations.\href{https://paperpile.com/c/crHGMz/PTc5}{\textsuperscript{7}}

\section{Conclusion}
This study demonstrates the feasibility of using LLMs for extracting clinically relevant data from unstructured social media content. The approach was cost-effective and computationally efficient, enabling a research team to conduct a study faster and with fewer resources while maintaining good accuracy and reliability.

\newpage
\bibliographystyle{unsrt}

\end{document}